\documentclass{article}
\usepackage{amsmath}
\usepackage{graphicx}
\usepackage[numbers]{natbib}
\usepackage{wrapfig}
\usepackage{url}
\usepackage{bm}
\usepackage[preprint]{neurips_2026}
\makeatletter
\renewcommand{\@noticestring}{}
\makeatother
\usepackage[utf8]{inputenc} 
\usepackage[T1]{fontenc}    
\usepackage{hyperref}       
\usepackage{url}            
\usepackage{booktabs}       
\usepackage{amsfonts}       
\usepackage{nicefrac}       
\usepackage{microtype}      
\usepackage{xcolor}         

\title{Motif-Mamba: network motif improved mamba for long-range sequence modeling}

\author{%
  Chonghe Hao \quad
  Yue Sun \quad
  Jian Zhang \quad
  Yansong Wang \\
  Wangzi Yao \quad
  Yunjie Yao \quad
  Tielin Zhang \\
  Center for Excellence in Brain Science and Intelligence Technology,\\
  Chinese Academy of Sciences \\
  \texttt{haoch2025@ion.ac.cn}
}

\begin{document}

\maketitle

\begin{abstract}
Efficient long-sequence modeling remains a central challenge for large language models, as self-attention scales quadratically with sequence length. Mamba offers a linear-time alternative through selective state space recurrence, but its predominantly diagonal state transitions restrict explicit interactions among state dimensions. We propose Motif-Mamba, a structured state space model that augments Mamba with a motif-constrained low-rank recurrent pathway. Inspired by the dynamics of three-node network motifs, the proposed pathway projects hidden states into a compact dynamical subspace, imposes motif-guided interactions, and maps the resulting dynamics back to the original state space. This design enhances cross-dimensional communication while preserving the linear-time recurrent structure of Mamba. Experiments on long-sequence extrapolation, language modeling benchmarks, and brain--computer interface decoding show consistent improvements over Mamba backbones, suggesting that motif-guided low-rank dynamics provide an effective structural prior for long-range sequence modeling.
\end{abstract}

\section{Introduction}

\begin{figure}[t]
    \centering
    \includegraphics[width=0.95\linewidth]{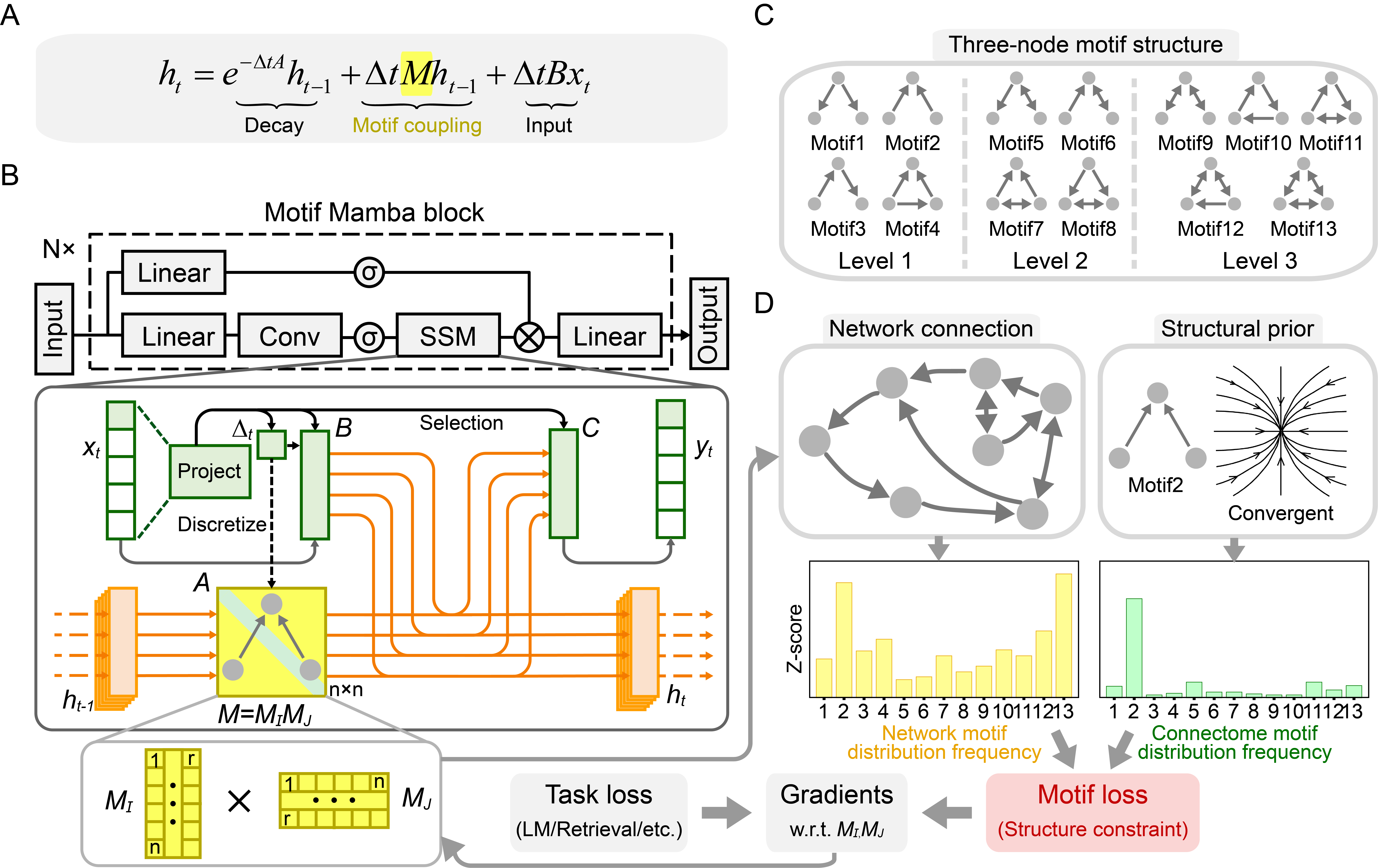}
    \caption{
    Overview of Motif-Mamba.
    (A) Motif-Mamba adds a motif-constrained coupling term to the standard Mamba state update.
    (B) Motif-Mamba architecture. The input is processed through the selective SSM pathway, while the motif-constrained low-rank matrix $M = M_I M_J$ provides an additional recurrent coupling path for efficient cross-dimensional state interaction.
    (C) Three-node motifs are used to describe local connectivity patterns.
    (D) Motif statistics from the learned matrix are compared with target motif statistics to form the motif loss, which jointly optimizes the structural constraint and the task objective.
    }
    \label{fig:motif_mamba_overview}
\end{figure}

Large language models have achieved remarkable progress in natural language understanding, text generation, code reasoning, and multimodal modeling \citep{brown2020language}. As their application scenarios expand, effective long-sequence processing becomes increasingly important for tasks such as long-document understanding, multi-turn dialogue, code repository analysis, knowledge retrieval, and scientific time-series modeling \citep{beltagy2020longformer}. However, most existing large language models are based on the Transformer architecture, whose self-attention mechanism explicitly computes pairwise token interactions and incurs quadratic computational and memory costs with respect to sequence length \citep{vaswani2017attention}. Although prior work has explored sparse attention, long-document Transformers, and memory-efficient attention to alleviate this bottleneck \citep{child2019generating,zaheer2020bigbird,dao2022flashattention}, efficient long-sequence modeling remains a central challenge.

State space models provide an efficient alternative to attention by propagating information through recurrent state updates \citep{gu2022s4}. Related recurrent and retention-based architectures have also been explored for efficient sequence modeling \citep{peng2023rwkv,sun2023retentive}. Mamba further introduces input-dependent selection and enables linear-time sequence modeling, making it attractive for long-context modeling \citep{gu2023mamba}. However, its diagonal or nearly diagonal state transitions limit direct interactions among state dimensions, restricting cross-dimensional dynamics and long-range dependency modeling.

Network motifs are recurring local connectivity patterns in complex networks and have been widely studied as functional building blocks of structured systems \citep{milo2002network}. Recent work shows that 13 directed three-node motifs can be grouped into a three-level dynamical hierarchy, where different motif groups correspond to different stability and flexibility regimes \citep{functional2026building,yao2026critical,prill2005dynamic}. This suggests that motifs are not only static graph patterns, but also carry interpretable dynamical properties. Moreover, motif-based analysis provides a compact way to link local connectivity patterns with higher-level network function \citep{alon2007network}. Motivated by this view, we introduce motif-based structural priors into Mamba, aiming to guide hidden state interactions with more organized and interpretable dynamical patterns.

To efficiently incorporate this motif prior, we propose Motif-Mamba, a structured state space model that implements motif-constrained state interaction through a low-rank dynamical pathway. Specifically, Motif-Mamba projects the high-dimensional hidden state into a compact low-dimensional space, performs motif-constrained interaction in that space, and maps the result back to the original state space. This design enables structured cross-dimensional interaction without using a dense transition matrix, while remaining consistent with prior evidence that low-rank structures can reveal compact neural dynamics \citep{sainath2013lowrank,mastrogiuseppe2018linking}.

By introducing motif-constrained low-rank dynamics, Motif-Mamba encourages more organized state interactions and turns hidden state evolution into a more structured gray-box dynamical process. We evaluate Motif-Mamba on long-sequence extrapolation, standard language model benchmarks, ablation studies, and brain--computer interface decoding tasks. The results show that Motif-Mamba improves long-range dependency modeling while preserving the computational efficiency of Mamba.

Our primary contributions can be summarized as follows:
\begin{itemize}
    \item We propose Motif-Mamba, a motif-constrained state space model that enhances Mamba for efficient long-sequence modeling.

    \item We design a low-rank recurrent pathway that embeds motif-guided state interactions into Mamba while preserving linear-time sequence modeling.

    \item We evaluate Motif-Mamba on long-sequence extrapolation, standard language model benchmarks and brain--computer interface decoding, demonstrating consistent improvements over Mamba and other baselines.

    \item We analyze different motif constraints from the dynamics of motif and study how motifs with different dynamical properties affect hidden-state evolution.
\end{itemize}

\section{Method}

\subsection{Preliminaries: Selective State Space Models}

State space models describe sequence dynamics through recurrent hidden state updates and have been widely used for efficient long-range sequence modeling \citep{gu2022s4,gu2023mamba}. Given an input sequence $x(t)$, a continuous-time state space model can be written as
\begin{equation}
    \frac{d h(t)}{dt} = -A(t) h(t) + B(t) x(t),
\end{equation}
where $h(t)$ denotes the hidden state and $A(t) \in \mathbb{R}^{n \times n}$ is the general state transition matrix. After discretization, the state update can be expressed as
\begin{equation}
    h_t = \bar{A}_t h_{t-1} + \bar{B}_t x_t,
\end{equation}
where $\bar{A}_t$ and $\bar{B}_t$ denote the discretized transition and input terms respectively.

Mamba extends classical state space models with a selective mechanism where several parameters are dynamically generated from the input. In particular, the input-dependent step size $\Delta_t$, input projection $B_t$, and output projection $C_t$ allow the model to selectively preserve or forget information according to the current token. A simplified Mamba state update can be written as
\begin{equation}
    h_t = \exp(-\Delta_t A_t) h_{t-1} + \Delta_t B_t x_t .
\end{equation}

This recurrent formulation enables linear-time sequence processing. However, to maintain computational efficiency, $A$ is usually parameterized as a diagonal matrix in practical SSM architectures. As a result, different state dimensions lack explicit coupling and tend to evolve independently. Although efficient, this structure limits the ability of the model to represent complex high-dimensional dynamics.

\subsection{Low-Rank Recurrent Coupling}

Motif-Mamba addresses this limitation by introducing a low-rank recurrent interaction pathway into the selective SSM. Let $M_I \in \mathbb{R}^{n \times r}$ and $M_J \in \mathbb{R}^{r \times n}$ denote the adapter matrices where $n$ is the hidden-state dimension and $r \ll n$ is the low-rank dimension. We consider the continuous-time SSM with this low-rank structure defined by the following differential equation

\begin{equation}
    \frac{d h(t)}{dt}
    =
    -Ah(t)
    +
    M_I M_J h(t)
    +
    B(t)x(t).
    \label{eq:lowrank_continuous_ssm}
\end{equation}

In this formulation, the term $M_I M_J h(t)$ introduces recurrent interaction among state dimensions through a compact low-dimensional space. This term is evaluated by first projecting the state to a latent state $z(t) = M_J h(t)$ and then mapping it back via $M_I z(t)$. Consequently, $z(t)$ serves as a compact latent state that drives the recurrent interaction.

\paragraph{Lemma 1. Exponential Euler discretization with low-rank interaction.}
Applying an exponential Euler discretization to the continuous-time dynamics in Eq.~\ref{eq:lowrank_continuous_ssm} leads to the following discrete-time approximation
\begin{equation}
    h_t
    =
    e^{-\Delta_t A}h_{t-1}
    +
    \Delta_t M_I M_J h_{t-1}
    +
    \Delta_t B_t x_t .
    \label{eq:motif_mamba_update}
\end{equation}
The detailed derivation is provided in Appendix~\ref{app:proof_lowrank_euler}. This formulation shows that the low-rank term acts as a recurrent driving component inside the SSM dynamics.

The update in Eq.~\ref{eq:motif_mamba_update} possesses several key properties. First, $M_I M_J$ is a square recurrent coupling matrix with rank at most $r$ allowing it to encode structured state interactions efficiently. Second, the product $M_I M_J h_{t-1}$ is evaluated as $M_I (M_J h_{t-1})$ which introduces cross-state coupling within a compact latent manifold. Third, while the diagonal self-decay term $e^{-\Delta_t A}h_{t-1}$ is preserved, the low-rank term $\Delta_t M_I M_J h_{t-1}$ injects structured cross-state information flow. Motif-Mamba thus introduces explicit recurrent interaction into Mamba while maintaining the efficient diagonal SSM backbone.

\subsubsection{Complexity Analysis}

The low-rank recurrent pathway introduces cross-dimensional state interaction without the computational burden of a dense transition matrix. A dense recurrent coupling $W h_{t-1}$ with $W \in \mathbb{R}^{n \times n}$ requires $O(n^2)$ computation per token. In contrast, Motif-Mamba computes $M_I (M_J h_{t-1})$ with a complexity of $O(nr)$. Since $r \ll n$, the low-rank coupling is significantly cheaper. For a sequence of length $L$, the recurrent complexity comparison is as follows

\begin{equation}
\left\{
\begin{aligned}
    \text{Standard Mamba:} \quad & O(Ln), \\
    \text{Motif-Mamba:} \quad & O(Ln) + O(Lnr), \\
    \text{Dense recurrent coupling:} \quad & O(Ln^2).
\end{aligned}
\right.
\end{equation}

Thus, Motif-Mamba adds explicit cross-state interaction while preserving the linear-time advantage of selective SSMs when $r$ is small.

\subsubsection{Recovering Hidden Dynamics from Latent States}

The latent state $z_t = M_J h_t$ defines the low-rank recurrent component $M_I z_t$ that enters the state update. Because this compact variable drives the cross-dimensional interaction, analyzing the system within this latent space provides two distinct analytical advantages.

First, $z_t$ establishes a rigorous mathematical basis for recovering the high-dimensional hidden-state trajectory. The following reconstruction property explains how the temporal evolution of $z_t$ captures the structural information of the complete system.

\paragraph{Lemma 2. Reconstruction from latent states and inputs.}
Consider the discrete Motif-Mamba recurrence where $z_{t-1} = M_J h_{t-1}$. Given the initial state $h_0$, the history of latent states $\{z_{\tau-1}\}_{\tau=1}^{t}$, the input history $\{x_\tau\}_{\tau=1}^{t}$ and the transition parameters $\{A_\tau, \Delta_\tau, B_\tau\}_{\tau=1}^{t}$, the hidden state $h_t$ is determined by
\begin{equation}
    h_t
    =
    \Gamma(t,0)h_0
    +
    \sum_{\tau=1}^{t}
    \Gamma(t,\tau)
    \left(
    \Delta_\tau M_I z_{\tau-1}
    +
    \Delta_\tau B_t x_\tau
    \right),
\end{equation}
where $\Gamma(t,\tau) = \prod_{j=\tau+1}^{t} e^{-\Delta_j A_j}$ and $\Gamma(t,t) = I$.
The proof is provided in Appendix~\ref{app:proof_z_reconstruction}.

This lemma mathematically guarantees that the sequence of $z_t$ along with the input history captures the complete recurrent contribution mediated by the low-rank pathway. 

Second, the low-dimensional $z_t$ provides an intuitive coordinate system for investigating the population dynamics of the system. Because $z_t$ directly defines the recurrent update, we can characterize the hidden dynamics by projecting the state evolution onto the $z$-plane. We define the projected flow field as
\begin{equation}
    v(z) = M_J A M_J^{+} z - z,
\end{equation}
where $M_J^{+}$ denotes the pseudoinverse of $M_J$ used to lift the latent state back to the hidden space. Analyzing this flow field reveals how motif constraints shape the collective state evolution across internal channels. Together, exact reconstruction and flow visualization establish the theoretical foundation for our experimental analysis.

\subsection{Motif-Constrained Structural Prior}

Network motifs act as functional building blocks that govern the stability and flexibility of recurrent interactions. To guide the hidden state evolution toward these interpretable dynamical regimes, we treat the low-rank coupling matrix $M = M_I M_J$ as a structurally constrained optimization object.

Calculating motif frequencies relies on a discrete adjacency structure which presents a challenge for backpropagation during gradient optimization. Following previous motif-guided optimization strategies \citep{functional2026building,yao2026critical}, we overcome this by introducing a continuous relaxation and substituting the binary adjacency matrix with a differentiable proxy $\tilde{M} \approx \sigma(\beta(M \odot M - \theta))$ where $\odot$ denotes the Hadamard product, $\sigma(\cdot)$ is the sigmoid function and $\beta$ is a temperature parameter controlling the sharpness of the threshold. Using this differentiable proxy, we extract the learned motif frequency vector
\begin{equation}
    \mathbf{m}_M = \mathcal{M}(\tilde{M}),
\end{equation}
where $\mathcal{M}(\cdot)$ represents the matrix-based motif extraction operator detailed in Appendix~\ref{app:motif_counting}. Given a target connectivity matrix representing the desired dynamical regime, we similarly compute its target motif frequency vector $\mathbf{m}_{\mathrm{target}}$.

We measure the structural alignment between the learned and target motif profiles using the coefficient of determination
\begin{equation}
    R^2
    =
    1 -
    \frac{
    \sum_{k=1}^{K}
    \left(
    m_{M,k} - m_{\mathrm{target},k}
    \right)^2
    }{
    \sum_{k=1}^{K}
    \left(
    m_{\mathrm{target},k} - \bar{m}_{\mathrm{target}}
    \right)^2
    + \epsilon
    },
\end{equation}
where $K$ denotes the number of selected motif categories and $\epsilon$ provides numerical stability. The motif loss is then defined as the deviation from perfect alignment
\begin{equation}
    \mathcal{L}_{\mathrm{motif}} = 1 - R^2 .
\end{equation}
This differentiable formulation enables targeted motif distributions to be reliably embedded into the network structure.

During training, the final objective function combines the standard task loss $\mathcal{L}_{\mathrm{task}}$ with the motif-based structural regularization
\begin{equation}
    \mathcal{L}
    =
    \mathcal{L}_{\mathrm{task}}
    +
    \lambda \mathcal{L}_{\mathrm{motif}},
\end{equation}
where $\lambda$ controls the strength of the structural constraint. This joint optimization ensures that the learned representation satisfies the specific sequence modeling requirements while maintaining a motif-guided dynamical organization.

\section{Experiments}
\subsection{Long-Sequence Extrapolation}

To evaluate long-sequence extrapolation, we use the induction heads task, a synthetic associative recall task that requires the model to remember a previous cue--target association and copy the corresponding continuation when the cue reappears. This provides a controlled setting for testing long-context memory.

Unless otherwise specified, Motif-Mamba denotes the Motif-2-constrained variant. This selection is derived from the three level motif hierarchy where different three node configurations correspond to distinct structural and dynamical regimes \citep{functional2026building,prill2005dynamic}. The topological features of Motif 2 are characterized by two nodes converging toward a single target without any reciprocal edges or closed loops. In contrast Motif 12 incorporates reciprocal connections and forms a cyclic loop structure. Previous dynamical analyses indicate that these topological differences fundamentally alter the stability of the recurrent interaction \citep{functional2026building, yao2026critical, prill2005dynamic}. The reciprocal edges and cycles in Motif 12 create feedback loops that are harder to stabilize than the convergent structure of Motif 2. This structural complexity makes Motif 12 inherently more difficult to optimize during training.


All models are trained with sequence length 256 and vocabulary size 16, and are evaluated on sequence lengths from 64 to 16384 tokens. We compare Motif-Mamba with representative sequence modeling baselines, including LSTM \citep{hochreiter1997long}, Transformer \citep{vaswani2017attention}, RWKV \citep{peng2023rwkv}, and standard Mamba \citep{gu2023mamba}. We use the same data generation process, optimizer, batch size, learning rate schedule, and training steps across models. The metric is next-token prediction accuracy at induction target positions.

\begin{figure}[t]
    \centering

    \begin{minipage}{0.48\linewidth}
        \centering
        \includegraphics[width=\linewidth]{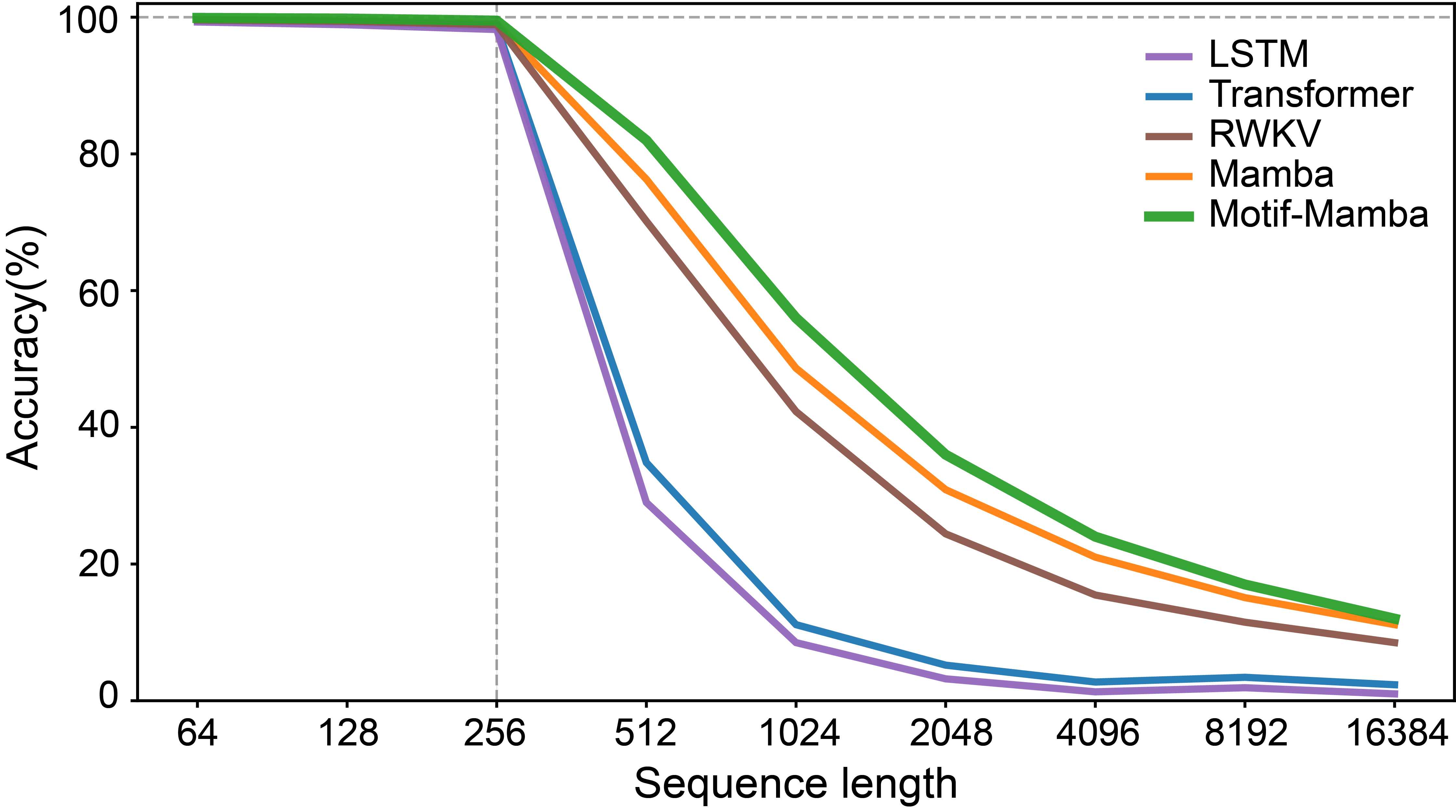}
        \caption{
        Long-sequence extrapolation on the induction heads task.
        Models are trained with 256 tokens and tested from 64 to 16384 tokens.
        }
        \label{fig:long_extrapolation}
    \end{minipage}
    \hfill
    \begin{minipage}{0.48\linewidth}
        \centering
        \includegraphics[width=\linewidth]{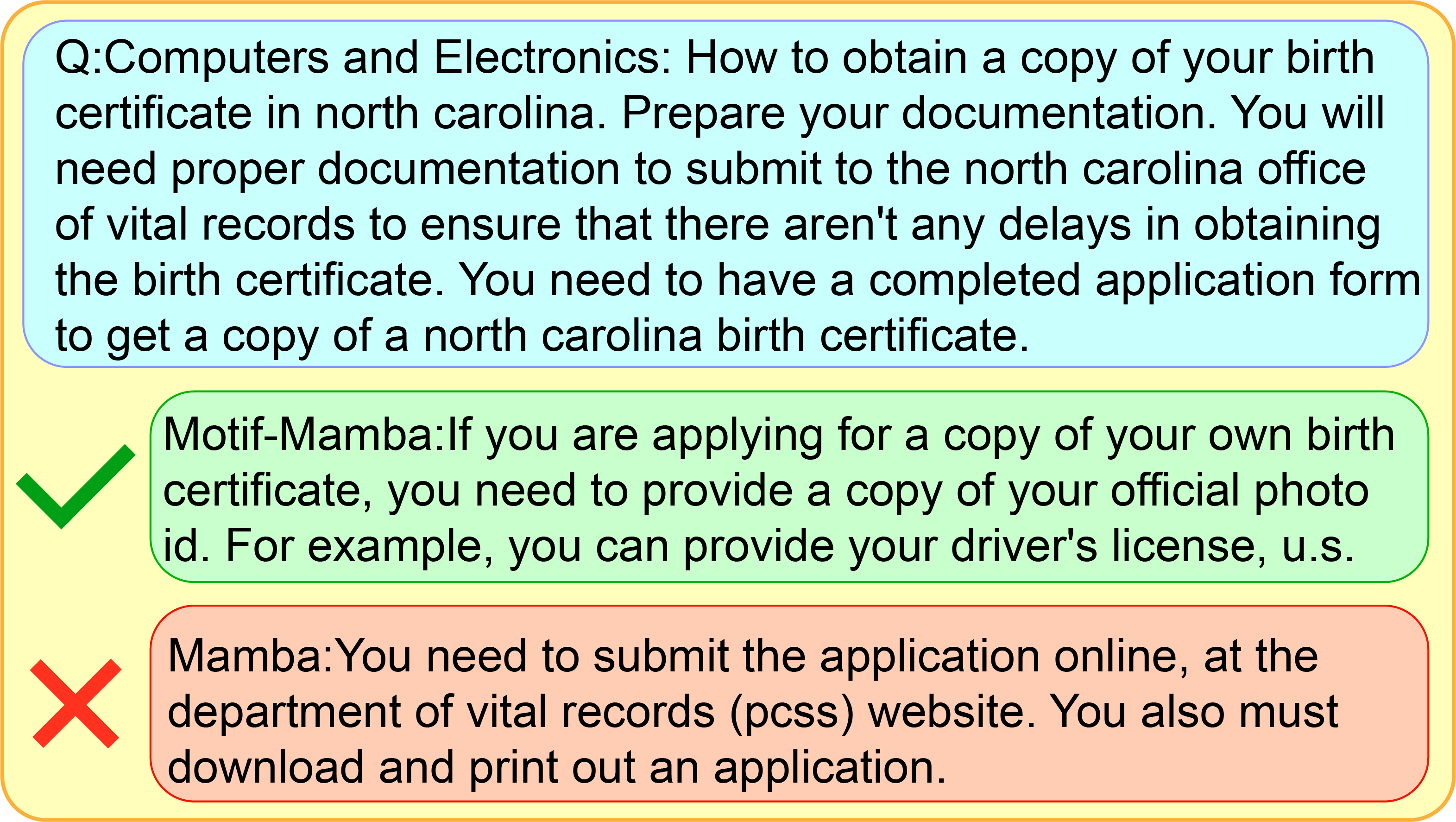}
        \caption{
        Illustrative example of Motif-Mamba performing a long-context question answering task.
        }
        \label{fig:qa}
    \end{minipage}

\end{figure}

Figure~\ref{fig:long_extrapolation} shows that Motif-Mamba maintains stronger extrapolation performance as the test length increases. Compared with standard Mamba and other baselines, Motif-Mamba is more stable beyond the training context, suggesting that motif-constrained low-rank dynamics help preserve and retrieve task-relevant information over long sequences. The numerical results corresponding to the curves in Figure~\ref{fig:long_extrapolation} are provided in Table~\ref{tab:induction_heads_extrapolation} as a complementary quantitative summary.

\subsection{Language Model Evaluation}

We further evaluate the general language modeling ability of Motif-Mamba on standard benchmarks from the lm-evaluation-harness \citep{evalharness}, including LAMBADA \citep{paperno2016lambada}, HellaSwag \citep{zellers2019hellaswag}, PIQA \citep{bisk2020piqa}, ARC-E and ARC-C \citep{clark2018arc}, and WinoGrande \citep{sakaguchi2021winogrande}. These tasks cover language modeling, commonsense reasoning, physical reasoning, and multiple-choice question answering, providing a broad evaluation of model capability beyond the synthetic long-sequence setting. Figure~\ref{fig:qa} shows a qualitative long-context QA example, where Motif-Mamba retrieves the relevant evidence and produces the correct answer, while standard Mamba fails to preserve the task-relevant information.

\begin{table}[h!]
\caption{
Language model evaluation results on standard benchmarks (\%).
}
\label{tab:LM evaluation}
\centering
\scriptsize
\begin{tabular}{lllcccccccc}
\toprule
\textbf{Model} &  LAMB. & LAMB. & HellaS. & PIQA & Arc-E & Arc-C & WinoGr. & Average \\
 & PPL $\downarrow$ & ACC $\uparrow$ & ACC $\uparrow$ & ACC $\uparrow$ & ACC $\uparrow$ & ACC $\uparrow$ & ACC $\uparrow$ & ACC $\uparrow$ \\
\midrule

Hybrid H3-130M~\citep{fu2023h3} & 89.48 & 25.8 & 31.7 & 64.2 & 44.4 & 24.2 & 50.5 & 40.1 \\
Pythia-160M~\citep{biderman2023pythia} & 38.10 & 33.0 & 30.2 & 61.4 & 43.2 & 24.1 & 51.9 & 40.6 \\
Mamba-130M~\citep{gu2023mamba} & 16.07 & 44.3 & 35.3 & 64.5 & 48.0 & 24.3 & 51.9 & 44.7 \\
\textbf{Motif-Mamba-130M} & \textbf{15.47} & \textbf{44.7} & \textbf{35.7} & \textbf{64.7} & \textbf{48.5} & \textbf{24.6} & \textbf{52.4} & \textbf{45.1} \\
\midrule

Hybrid H3-360M~\citep{fu2023h3} & 12.58 & 48.0 & 41.5 & 68.1 & 51.4 & 24.7 & 54.1 & 48.0 \\
Pythia-410M~\citep{biderman2023pythia} & 10.84 & 51.4 & 40.5 & 66.9 & 52.1 & 24.6 & 53.8 & 48.2 \\
Mamba-370M~\citep{gu2023mamba} & 8.14 & 55.6 & 46.5 & 69.5 & 55.1 & 28.0 & 55.3 & 51.6 \\
\textbf{Motif-Mamba-370M} & \textbf{7.95} & \textbf{55.9} & \textbf{47.1} & \textbf{70.0} & \textbf{55.6} & \textbf{28.7} & \textbf{56.3} & \textbf{52.3} \\
\midrule

Pythia-1B~\citep{biderman2023pythia} & 7.92 & 56.1 & 47.2 & 70.7 & 57.0 & 27.1 & 53.5 & 51.9 \\
Mamba-790M~\citep{gu2023mamba} & 6.02 & 62.7 & 55.1 & 72.1 & 61.2 & 29.5 & 56.1 & 56.1 \\
\textbf{Motif-Mamba-790M} & \textbf{5.88} & \textbf{62.8} & \textbf{55.9} & \textbf{72.6} & \textbf{61.5} & \textbf{29.6} & \textbf{57.1} & \textbf{56.6} \\
\midrule

GPT-Neo 1.3B~\citep{black2021gptneo} & 7.50 & 57.2 & 48.9 & 71.1 & 56.2 & 25.9 & 54.9 & 52.4 \\
Hybrid H3-1.3B~\citep{fu2023h3} & 11.25 & 49.6 & 52.6 & 71.3 & 59.2 & 28.1 & 56.9 & 53.0 \\
OPT-1.3B~\citep{zhang2022opt} & 6.64 & 58.0 & 53.7 & 72.4 & 56.7 & 29.6 & 59.5 & 55.0 \\
Pythia-1.4B~\citep{biderman2023pythia} & 6.08 & 61.7 & 52.1 & 71.0 & 60.5 & 28.5 & 57.2 & 55.2 \\
RWKV-1.5B~\citep{peng2023rwkv} & 7.04 & 56.4 & 52.5 & 72.4 & 60.5 & 29.4 & 54.5 & 54.3 \\
Mamba-1.4B~\citep{gu2023mamba} & 5.04 & 64.9 & 59.1 & 74.2 & 65.5 & 32.8 & 61.5 & 59.7 \\
\textbf{Motif-Mamba-1.4B} & \textbf{4.82} & \textbf{65.5} & \textbf{60.5} & \textbf{75.1} & \textbf{66.8} & \textbf{33.9} & \textbf{62.7} & \textbf{60.8} \\

\bottomrule
\end{tabular}
\end{table}

We compare Motif-Mamba with standard Mamba and other representative baseline models of similar scale. The evaluation metrics include LAMBADA perplexity and accuracy-based scores for the remaining tasks. Unless otherwise specified, all models are evaluated under the same protocol. In the main experiments, Motif-Mamba uses the default Motif-2 constraint.

Table~\ref{tab:LM evaluation} reports the language model evaluation results. Motif-Mamba consistently improves over the corresponding Mamba backbone across different model sizes. The gains are observed in both LAMBADA perplexity and average accuracy, suggesting that motif-constrained low-rank state interaction not only benefits synthetic long-sequence extrapolation, but also improves general language modeling and reasoning performance.

\subsection{BCI Decoding}

We evaluate Motif-Mamba on brain--computer interface decoding tasks using the JangoBCI dataset. This dataset contains neural and behavioral recordings from a non-human primate performing an isometric wrist task, and has been used for evaluating long-term BCI decoding and neural alignment methods \citep{ma2023adversarial}. Neural signals are naturally long temporal sequences and often contain dependencies across both time and recording channels. Therefore, BCI decoding provides a useful testbed for evaluating whether the proposed structured low-rank dynamics can generalize to non-text sequential data.

\begin{figure}
    \centering
    \includegraphics[width=1\linewidth]{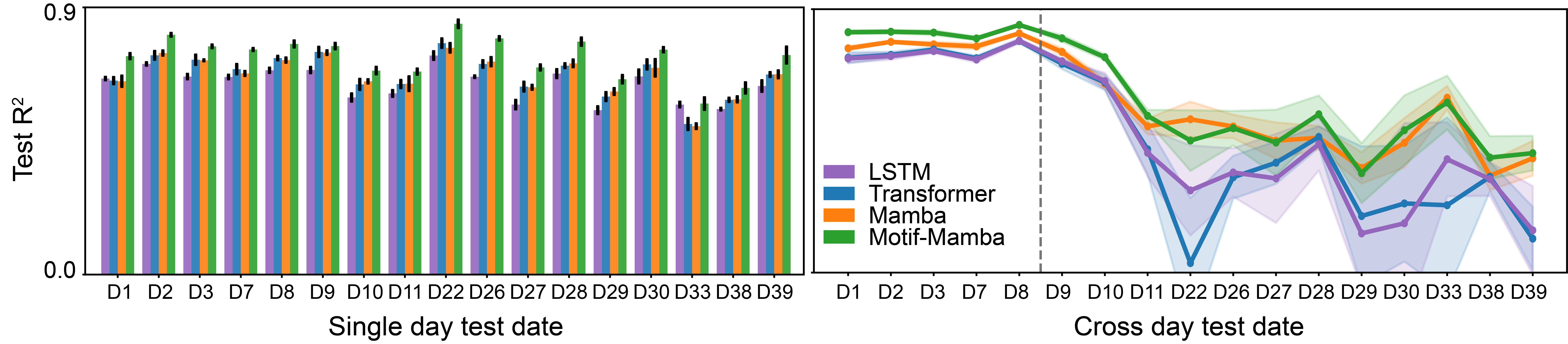}
    \caption{
    BCI decoding results on the JangoBCI dataset.
    The figure compares Motif-Mamba with baseline sequence models on long temporal neural signals collected across recording sessions from Day 1 (D1) to Day 39 (D39).
    The first panel reports single-day training and single-day testing performance, where error bars denote std.
    The second panel reports cross-day generalization, where data before the dashed line are used for training and data after the dashed line are used only for evaluation; shaded regions denote std ($n=5$).
    }
    \label{fig:bci_decoding}
\end{figure}

Given neural signal inputs, the model is trained to predict the corresponding behavioral or task-related outputs. We compare Motif-Mamba with standard Mamba and commonly used sequence models, including LSTM and Transformer. All models are trained and evaluated using the same data split, preprocessing pipeline, and evaluation protocol.

Figure~\ref{fig:bci_decoding} shows the decoding performance of different models. Motif-Mamba achieves better decoding performance than the baseline models, suggesting that motif-constrained low-rank dynamics can help capture long-range temporal dependencies in neural time series. These results provide additional evidence that Motif-Mamba is not limited to language modeling and may also be useful for complex temporal signal decoding.


\subsection{The Role of Motif-Constrained Low-Rank Dynamics}

We first conduct ablation studies to examine the role of motif constraints in Motif-Mamba. The goal is to separate the effect of the additional low-rank recurrent pathway from the structural constraint. We compare four variants including the original Mamba, Motifblank-Mamba with unconstrained low-rank coupling, and the constrained variants Motif2-Mamba and Motif12-Mamba.

\begin{table}[h!]
\caption{
Ablation results on standard benchmarks (\%).
}
\label{tab:ablation}
\centering
\scriptsize
\begin{tabular}{lllcccccccc}
\toprule
\textbf{Model} & LAMB. & LAMB. & HellaS. & PIQA & Arc-E & Arc-C & WinoGr. & Average \\
 & PPL $\downarrow$ & ACC $\uparrow$ & ACC $\uparrow$ & ACC $\uparrow$ & ACC $\uparrow$ & ACC $\uparrow$ & ACC $\uparrow$ & ACC $\uparrow$ \\
\midrule

\textbf{Mamba-130M} & 16.07 & 44.3 & 35.3 & 64.5 & 48.0 & 24.3 & 51.9 & 44.7 \\
\midrule

\textbf{Motifblank-Mamba-130M} & 15.79 & 44.3 & 35.2 & 64.5 & 48.0 & 24.5 & 52.3 & 44.8 \\
\textbf{Motif2-Mamba-130M} & \textbf{15.47} & \textbf{44.7} & \textbf{35.7} & \textbf{64.7} & \textbf{48.5} & \textbf{24.6} & \textbf{52.4} & \textbf{45.1} \\
\textbf{Motif12-Mamba-130M} & 15.88 & 44.2 & 35.5 & 64.6 & 48.4 & 24.5 & 51.9 & 44.9 \\

\bottomrule
\end{tabular}
\end{table}

Table~\ref{tab:ablation} reports the ablation results on standard language model benchmarks. Compared with Mamba-130M, Motifblank-Mamba-130M shows almost no improvement as the average accuracy only increases from 44.7\% to 44.8\%. This suggests that simply adding an unconstrained low-rank pathway is insufficient. In contrast, motif-constrained variants achieve clearer gains which indicates the importance of structural constraints. Among them, Motif2-Mamba-130M performs best and achieves the highest average accuracy of 45.1\%.

\begin{figure}[t]
    \centering
    \includegraphics[width=\linewidth]{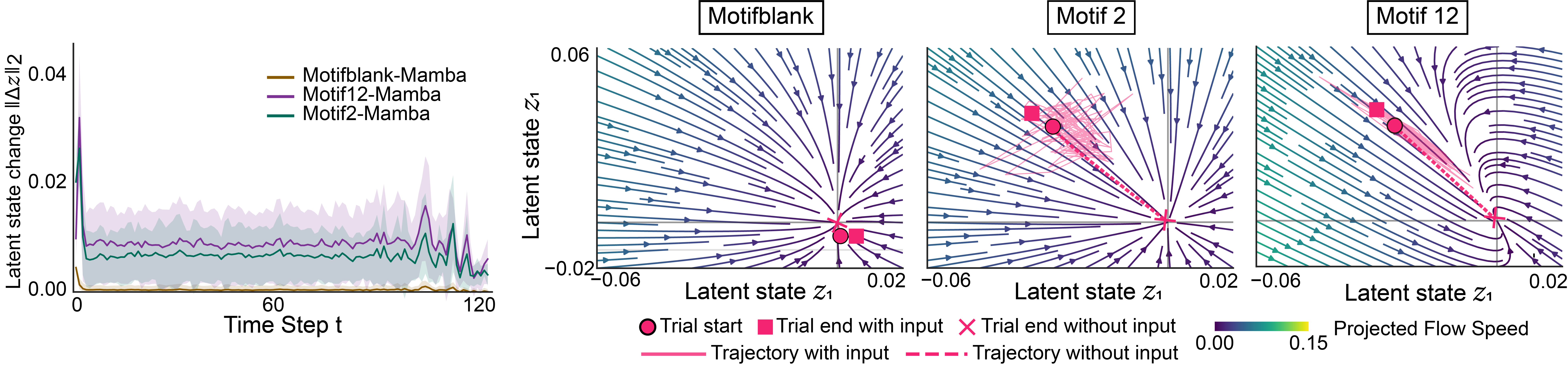}
    \caption{
    Motif constraints induce structured low-rank recurrent dynamics. 
    Projected flow fields and latent trajectories in the rank-2 $z$-plane for Motifblank, Motif2 and Motif12. 
    Solid pink lines denote empirical trajectories with external inputs while dashed pink lines denote autonomous continuations without input. 
    The background vector field indicates the projected autonomous flow $v(z)$ colored by its local speed.
    }
    \label{fig:latent_dynamics}
\end{figure}

To understand why these structural constraints improve performance we analyze the low-rank recurrent dynamics induced by the low-rank coupling. As established in Lemma 1 and Lemma 2 the latent state $z_t$ actively drives the recurrent interaction and provides a rigorous basis for state reconstruction. Consequently $z_t$ serves as a natural low-dimensional latent state for analyzing the induced population dynamics. 

Following the theoretical framework from the Method section, we constructed the projected autonomous flow fields $v(z)$ in the rank-2 $z$-plane (Fig.~\ref{fig:latent_dynamics}). Although the full high-dimensional hidden state $h$ may contain components in the null space of $M_J$, the continuations without input initialized from real trial states closely followed the local flow directions. This suggests that the low-dimensional field effectively captures the autonomous component of the low-rank recurrent interaction. Our analysis reveals that motif-constrained structures induce more pronounced low-rank recurrent interactions compared to the unconstrained Motifblank architecture. 
\begin{wrapfigure}{r}{0.42\textwidth}
    \centering
    \includegraphics[width=0.40\textwidth]{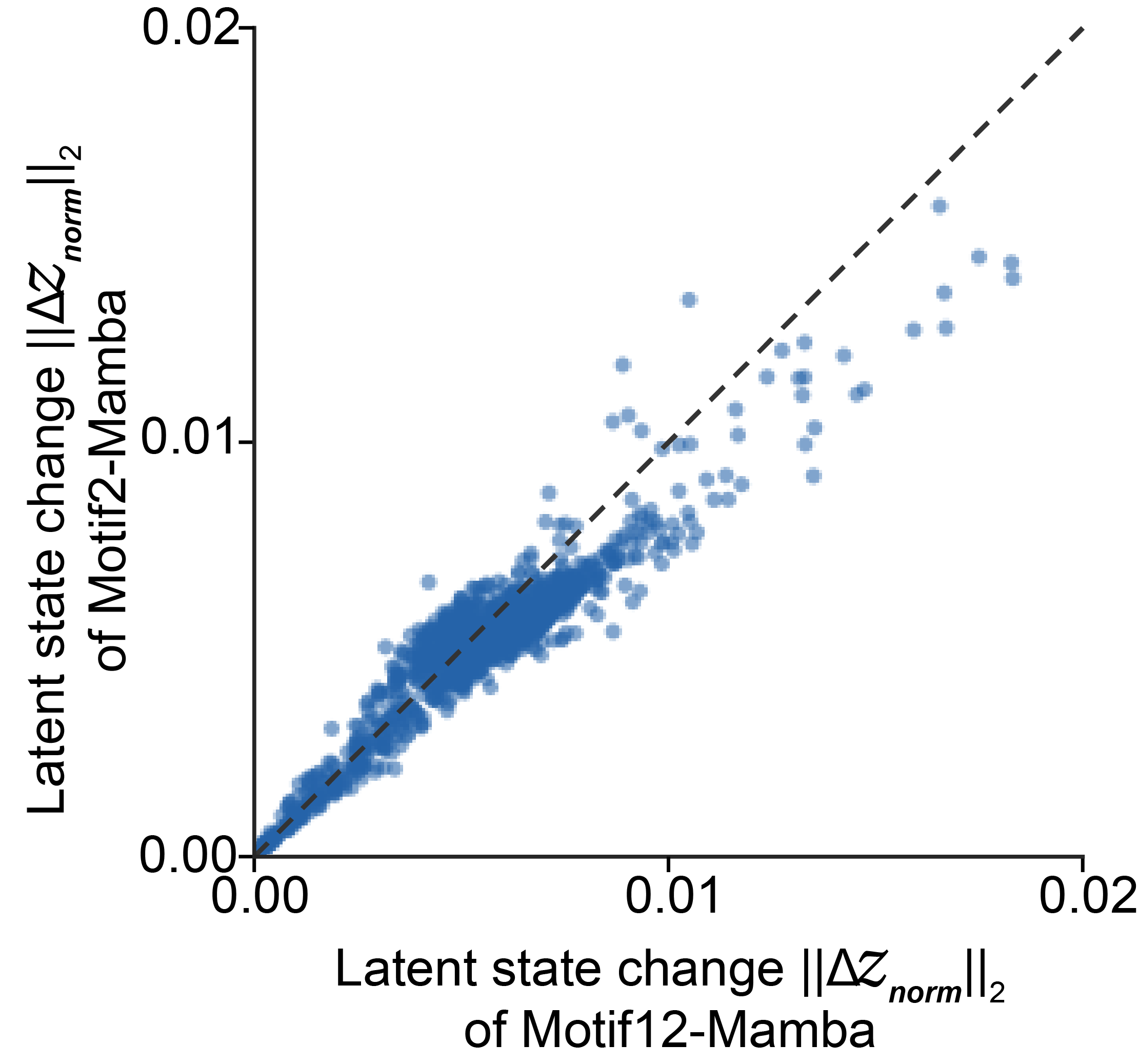}
    \caption{
    Population-level comparison of latent interaction magnitudes.
    Each point represents the average $\| \Delta z_{norm} \|_2$ over 200 trials for an individual layer internal channel ($n=1531$). 
    The coordinates compare the interaction strength of Motif12-Mamba (x-axis) against Motif2-Mamba (y-axis). 
    The dashed diagonal indicates the identity line where values are equal. 
    The heavy concentration of points below this line demonstrates that Motif12 consistently induces stronger low-rank recurrent interactions than Motif2.
    }
    \label{fig:latent_channel_scatter}
\end{wrapfigure}
In Motifblank, both the projected flow and empirical trajectories remain clustered near the origin. This indicates that the low-rank pathway fails to produce salient recurrent interactions in the latent space. In contrast, Motif2 and Motif12 exhibit well-defined $z$-plane trajectories and significantly larger flow magnitudes.

To quantify this structural displacement, we measured the $M_I$-normalized latent state change $\| \Delta z_{norm}(t) \|_2 = \| z_{norm}(t) - z_{norm}(t-1) \|_2$, where $z_{norm,a}(t) = \| M_{I,:,a} \|_2 (M_J)_a h_{t-1}$. Motif2 and Motif12 demonstrated substantially higher $\| \Delta z_{norm}(t) \|_2$ than Motifblank across 200 LAMBADA examples which confirms that motif constraints activate stronger and more structured low-rank recurrent interactions. Finally, we compared the interaction magnitudes of Motif2 and Motif12 across all layer 0 channels (Fig.~\ref{fig:latent_channel_scatter}). We calculated the average $\| \Delta z_{norm} \|_2$ over 200 LAMBADA examples for each channel. In the channel-wise scatter plot, the majority of data points fall below the identity line. This indicates that Motif12 generally induces larger latent state changes than Motif2. Such an observation suggests that Motif12 facilitates stronger low-rank recurrent interactions across the internal channel population. These findings align with previous motif hierarchy studies which suggested that different three-node motif classes are associated with distinct dynamical regimes \citep{functional2026building,prill2005dynamic}. Consequently, motif constraints do not merely activate the low-rank pathway, but precisely modulate the intensity and structure of the induced latent dynamics across channels.

\section{Conclusion and Discussion}

\paragraph{Conclusion.}
We introduced Motif-Mamba, a structured extension of Mamba designed to strengthen long-sequence modeling through motif-constrained state dynamics. The proposed motif-guided recurrent pathway enables efficient cross-dimensional state interaction while preserving the linear-time modeling advantage of Mamba. Experiments on long-sequence extrapolation, language model benchmarks, and BCI decoding show that Motif-Mamba improves both text and non-text temporal sequence modeling. Ablation results further demonstrate that the gains mainly come from the motif-based structural constraint rather than from unconstrained low-rank coupling alone. Overall, Motif-Mamba provides an efficient and structured approach for enhancing the dynamical expressiveness of state space models.

\paragraph{Discussion.}
The language modeling results suggest that motif-constrained dynamics are not limited to synthetic sequence tasks. Motif-Mamba improves the corresponding Mamba backbone across multiple model scales, indicating that the proposed structural prior does not simply overfit to a specific experimental setting. Language modeling requires coordinated integration of lexical, syntactic, and semantic information across hidden dimensions \citep{gu2023mamba,mastrogiuseppe2018linking}. The motif-constrained low-rank pathway provides a regularized interaction route among state dimensions, which may help organize hidden representations more effectively than an unconstrained low-rank coupling.

For long-sequence extrapolation, Motif-Mamba shows stronger stability when the test sequence length exceeds the training context. The induction heads task requires preserving cue--target associations over long temporal gaps, making it a controlled setting for evaluating long-context memory \citep{olsson2022context}. Standard Mamba benefits from selective state-space recurrence, but its mostly diagonal state evolution limits explicit communication among state channels \citep{gu2023mamba}. By adding a compact motif-guided dynamical pathway, Motif-Mamba improves the propagation of task-relevant information without relying on dense attention over the full context.

Among the tested motif constraints, Motif-2 achieves the best overall performance. One possible explanation is that Motif-2 belongs to the more stable group in the three-level motif hierarchy \citep{functional2026building}. Compared with motifs containing richer recurrent feedback, Motif-2 provides a simpler local interaction pattern, which may better support memory preservation, controlled state propagation, and reduced interference. The ablation results support this interpretation, as Motif-2 outperforms Motif-12, suggesting that more complex motif structures are not necessarily more beneficial for language and long-sequence tasks.

More broadly, Motif-Mamba provides a possible route for introducing motif-level structural priors inspired by biological connectivity into artificial sequence models. In this work, motif statistics are used as lightweight structural inspiration rather than a direct mechanistic model of biological neural circuits. Local organizational principles observed in biological connectivity may inspire more structured and interpretable mechanisms in neural architectures \citep{wang2020allen}. The motif prior makes part of the hidden-state evolution more analyzable, turning the recurrent update from a fully black-box process into a gray-box dynamical structure. Future work may extend this framework by assigning different motif distributions to different layers, tasks, or domains, allowing motif priors to adapt to task-specific dynamical requirements such as stable memory preservation, flexible updating, or robust temporal decoding.

\paragraph{Limitation.}

Several limitations remain. First, this work focuses on three-node motifs, which capture basic local connectivity patterns but cannot fully represent higher-order structural interactions. Extending motif-constrained modeling to larger subgraphs, such as four-node motifs or hierarchical motif compositions, may provide a richer structural prior. Second, the current study mainly uses fixed representative motif types from different stability levels. Although this design allows controlled analysis, it may not fully capture the diverse dynamical requirements of different tasks, where intermediate or mixed motif regimes could offer a better trade-off between stability and flexibility. Future work may therefore explore adaptive, layer-specific, or task-dependent motif priors to improve flexibility. Third, Motif-Mamba applies motif priors to state space dynamics, but how to define and compare motif-induced dynamics across different architectures remains an open question. A broader analysis of the relationship between motif hierarchy and model behavior across sequence modeling tasks may help build more efficient, structured, and interpretable state space models.

\bibliographystyle{unsrtnat}
\bibliography{references}





\clearpage
\appendix
\renewcommand{\thetable}{S\arabic{table}}
\setcounter{table}{0}
\renewcommand{\thefigure}{S\arabic{figure}}
\setcounter{figure}{0}

\renewcommand{\theHtable}{S\arabic{table}}
\renewcommand{\theHfigure}{S\arabic{figure}}
\section{Notation}
\label{app:notation}

\begin{table}[h!]
\centering
\caption{Notation used throughout the paper.}
\label{tab:notation}
\scriptsize
\setlength{\tabcolsep}{4pt}
\renewcommand{\arraystretch}{0.9}
\begin{tabular}{p{0.25\linewidth}p{0.65\linewidth}}
\toprule
\textbf{Symbol} & \textbf{Definition} \\
\midrule
\multicolumn{2}{l}{\textbf{State space model}} \\
$x_t$, $x(t)$ & Discrete and continuous input. \\
$h_t$, $h(t)$ & Discrete and continuous hidden state. \\
$A_t$, $A(t)$ & Discrete and continuous state transition matrix. \\
$B_t$, $B(t)$ & Discrete and continuous input projection matrix. \\
$C_t$ & Output projection matrix. \\
$\bar{A}_t$, $\bar{B}_t$ & Discretized transition and input terms. \\
$\Delta_t$ & Step size / selective timescale. \\
$n$, $L$ & Hidden-state dimension and sequence length. \\
\midrule
\multicolumn{2}{l}{\textbf{Low-rank motif pathway}} \\
$r$ & Low-rank dimension. \\
$M_I \in \mathbb{R}^{n \times r}$ & Latent-to-state low-rank factor. \\
$M_J \in \mathbb{R}^{r \times n}$ & State-to-latent low-rank factor. \\
$M=M_IM_J$ & Low-rank recurrent coupling matrix. \\
$z_t=M_Jh_t$ & Low-dimensional latent state. \\
$v(z)$ & Projected latent flow field. \\
$k_{\mathrm{motif}}$ & Motif coupling coefficient. \\
$\Delta z_{\mathrm{norm}}$ & Normalized latent-state displacement. \\
\midrule
\multicolumn{2}{l}{\textbf{Motif regularization}} \\
$\tilde{M}$ & Differentiable adjacency proxy. \\
$\sigma(\cdot)$ & Sigmoid function. \\
$\beta$, $\theta$ & Relaxation sharpness and threshold. \\
$\mathcal{M}(\cdot)$ & Motif extraction operator. \\
$\mathbf{m}_M$ & Learned motif frequency vector. \\
$\mathbf{m}_{\mathrm{target}}$ & Target motif frequency vector. \\
$m_{M,k}$, $m_{\mathrm{target},k}$ & Learned and target frequency of motif $k$. \\
$K$ & Number of motif categories. \\
$R^2$ & Motif-profile alignment score. \\
$\epsilon$ & Numerical stabilizer. \\
$\mathcal{L}_{\mathrm{task}}$ & Task loss. \\
$\mathcal{L}_{\mathrm{motif}}$ & Motif regularization loss. \\
$\mathcal{L}$ & Total training objective. \\
$\lambda$ & Motif regularization weight. \\
\midrule
\multicolumn{2}{l}{\textbf{Motif counting}} \\
$N$ & Number of graph nodes. \\
$i,j,k$ & Node indices. \\
$\widetilde{W}$ & Binary adjacency matrix. \\
$W$, $\bar{W}$ & Edge-present and edge-absent matrices. \\
$P$ & Off-diagonal mask. \\
$\ell$ & All-one vector. \\
$c_m$ & Count of the $m$-th directed three-node motif. \\
\midrule
\multicolumn{2}{l}{\textbf{Proofs and experiments}} \\
$\Gamma(t,\tau)$ & Transition product from step $\tau$ to step $t$. \\
$d_{\mathrm{state}}$, $d_{\mathrm{conv}}$ & Mamba state size and convolution width. \\
$n_{\mathrm{layer}}$ & Number of layers. \\
PPL, ACC, std & Perplexity, accuracy, and standard deviation. \\
\bottomrule
\end{tabular}
\end{table}

\section{Proofs and Derivations}

\subsection{Derivation of Lemma 1}
\label{app:proof_lowrank_euler}

We consider the continuous-time dynamics of the Motif-Mamba system defined by the state equation
\begin{equation}
    \dot h(\tau)
    =
    -A(\tau)h(\tau)
    +
    M_I M_J h(\tau)
    +
    B(\tau)x(\tau).
\end{equation}
For a discretization interval $[\tau_{t-1}, \tau_t]$ with step size $\Delta_t = \tau_t - \tau_{t-1}$, the solution follows the variation of constants formula
\begin{equation}
\begin{aligned}
h(\tau_t)
&=
\underbrace{
\exp\!\left(
-\int_{\tau_{t-1}}^{\tau_t} A(s) ds
\right)
h(\tau_{t-1})
}_{\text{self-decay term}}
\\
&\quad+
\underbrace{
\int_{\tau_{t-1}}^{\tau_t}
\exp\!\left(
-\int_{\tau}^{\tau_t} A(s) ds
\right)
\left(
M_I M_J h(\tau) + B(\tau)x(\tau)
\right)
d\tau
}_{\text{driving term}} .
\end{aligned}
\end{equation}
To derive the discrete-time update, we apply a piecewise constant approximation where the transition matrix $A(\tau) \approx A_t$ and the driving components $M_I M_J h(\tau) \approx M_I M_J h_{t-1}$ and $B(\tau)x(\tau) \approx B_t x_t$ remain constant over the interval. This assumption leads to the following integral approximation
\begin{equation}
\begin{aligned}
h_t
&\approx
e^{-\Delta_t A_t}h_{t-1}
+
\left(
\int_{\tau_{t-1}}^{\tau_t}
e^{-(\tau_t-\tau)A_t}
d\tau
\right)
\left(
M_I M_J h_{t-1} + B_t x_t
\right)
\\
&=
e^{-\Delta_t A_t}h_{t-1}
+
A_t^{-1}\left(I - e^{-\Delta_t A_t}\right)
\left(
M_I M_J h_{t-1} + B_t x_t
\right).
\end{aligned}
\end{equation}
By applying a first-order Taylor expansion $e^{-\Delta_t A_t} = I - \Delta_t A_t + O(\Delta_t^2)$ around $\Delta_t = 0$, the integral coefficient simplifies as follows
\begin{equation}
    A_t^{-1}\left(I - e^{-\Delta_t A_t}\right)
    =
    A_t^{-1}\left(\Delta_t A_t + O(\Delta_t^2)\right)
    =
    \Delta_t I + O(\Delta_t^2)
    \approx
    \Delta_t I .
\end{equation}
Substituting this approximation yields the discrete-time Motif-Mamba update without input derivative terms
\begin{equation}
    h_t
    \approx
    e^{-\Delta_t A_t}h_{t-1}
    +
    \Delta_t \left( M_I M_J h_{t-1} + B_t x_t \right) .
\end{equation}

\subsection{Proof of Lemma 2}
\label{app:proof_z_reconstruction}

Consider the discrete Motif-Mamba recurrence
\begin{equation}
    h_t
    =
    e^{-\Delta_t A_t}h_{t-1}
    +
    \Delta_t M_I z_{t-1}
    +
    \Delta_t B_t x_t .
\end{equation}
For \(0\leq \tau \leq t\), define the transition product
\begin{equation}
    \Gamma(t,\tau)
    =
    e^{-\Delta_t A_t}
    e^{-\Delta_{t-1}A_{t-1}}
    \cdots
    e^{-\Delta_{\tau+1}A_{\tau+1}},
\end{equation}
with the convention \(\Gamma(t,t)=I\). 
Equivalently, \(\Gamma(t,\tau)\) satisfies the recursion
\(\Gamma(t,\tau)=\Gamma(t,t-1)\Gamma(t-1,\tau)\), where
\(\Gamma(t,t-1)=e^{-\Delta_t A_t}\). 
Using this notation, the one-step recurrence can be written in terms of \(\Gamma\). Substituting the corresponding expression for \(h_{t-1}\) gives
\begin{equation}
\begin{aligned}
    h_t
    &=
    \Gamma(t,t-1)h_{t-1}
    +
    \Gamma(t,t)
    \left(
    \Delta_t M_Iz_{t-1}
    +
    \Delta_t B_tx_t
    \right)
    \\
&=
\Gamma(t,t-2)h_{t-2}
+
\Gamma(t,t-1)
\left(
\Delta_{t-1}M_Iz_{t-2}
+
\Delta_{t-1}B_{t-1}x_{t-1}
\right)
\\
&\quad+
\Gamma(t,t)
\left(
\Delta_t M_Iz_{t-1}
+
\Delta_t B_tx_t
\right).
\end{aligned}
\end{equation}
Repeating the same substitution from \(t\) back to \(0\) yields
\begin{equation}
    h_t
    =
    \Gamma(t,0)h_0
    +
    \sum_{\tau=1}^{t}
    \Gamma(t,\tau)
    \left(
    \Delta_\tau M_I z_{\tau-1}
    +
    \Delta_\tau B_\tau x_\tau
    \right).
\end{equation}
This proves the reconstruction formula. The expression shows that, once \(h_0\), the transition parameters, the input history, and the low-rank coordinate history \(\{z_{\tau-1}\}_{\tau=1}^{t}\) are given, the hidden state \(h_t\) is determined by the recurrence.


\section{Calculating the Frequency of Network Motifs}
\label{app:motif_counting}

Let $\widetilde{W}\in\{0,1\}^{N\times N}$ be the adjacency matrix, where $\widetilde{W}_{jk}=1$ indicates a directed connection from node $k$ to node $j$, and $\widetilde{W}_{jk}=0$ indicates its absence. Let $L=\mathbf{1}_{N,1}$ be an all-one vector. For any matrix $X\in\mathbb{R}^{N\times N}$,
\begin{equation}
    L^{\top}XL
    =
    \sum_{i,j} X_{ij}.
\end{equation}

To exclude self-loops, define
\begin{equation}
    P=\mathbf{1}_{N\times N}-I_{N\times N}.
\end{equation}
Then
\begin{equation}
    W_{ji}=\widetilde{W}_{ji}P_{ji},
    \qquad
    \bar{W}_{ji}=(P-\widetilde{W})_{ji},
\end{equation}
where \(W_{ji}\) denotes the presence of a directed edge and \(\bar{W}_{ji}\) denotes its absence.

For motif2, we consider the in-star pattern where two nodes \(j\) and \(k\) both project to node \(i\), while there are no reverse edges and no direct edges between \(j\) and \(k\). Its motif count is
\begin{equation}
    c_2(W)
    =
    \frac{1}{2}
    \sum_{i,j,k}
    W_{ij}
    \bar{W}_{ji}
    W_{ik}
    \bar{W}_{ki}
    \bar{W}_{kj}
    \bar{W}_{jk}.
\end{equation}

Equivalently, using matrix multiplication and Hadamard products, this count can be written as
\begin{equation}
\begin{aligned}
c_2(W)
&=
\frac{1}{2}
\sum_{i,j,k}
W_{ij}
\bar{W}_{ji}
W_{ik}
\bar{W}_{ki}
\bar{W}_{kj}
\bar{W}_{jk}
\\
&=
\frac{1}{2}
\sum_{i,j,k}
\widetilde{W}_{ij}
(P-\widetilde{W})_{ji}
\widetilde{W}_{ik}
(P-\widetilde{W})_{ki}
(P-\widetilde{W})_{kj}
(P-\widetilde{W})_{jk}
\\
&=
\frac{1}{2}
\sum_{i,j,k}
\widetilde{W}^{\top}_{ji}
(P-\widetilde{W})_{ji}
\widetilde{W}_{ik}
(P-\widetilde{W})^{\top}_{ik}
(P-\widetilde{W})^{\top}_{jk}
(P-\widetilde{W})_{jk}
\\
&=
\frac{1}{2}
\sum_{i,j,k}
\left[
\widetilde{W}^{\top}
\odot
(P-\widetilde{W})
\right]_{ji}
\left[
\widetilde{W}
\odot
(P-\widetilde{W})^{\top}
\right]_{ik}
\left[
(P-\widetilde{W})^{\top}
\odot
(P-\widetilde{W})
\right]_{jk}
\\
&=
\frac{1}{2}
\sum_{j,k}
\left(
\sum_i
\left[
\widetilde{W}^{\top}
\odot
(P-\widetilde{W})
\right]_{ji}
\left[
\widetilde{W}
\odot
(P-\widetilde{W})^{\top}
\right]_{ik}
\right)
\left[
(P-\widetilde{W})^{\top}
\odot
(P-\widetilde{W})
\right]_{jk}
\\
&=
\frac{1}{2}
\sum_{j,k}
\left[
\left(
\widetilde{W}^{\top}
\odot
(P-\widetilde{W})
\right)
\left(
\widetilde{W}
\odot
(P-\widetilde{W})^{\top}
\right)
\right]_{jk}
\left[
(P-\widetilde{W})^{\top}
\odot
(P-\widetilde{W})
\right]_{jk}
\\
&=
\frac{1}{2}
\sum_{j,k}
\left[
\left(
\widetilde{W}^{\top}
\odot
(P-\widetilde{W})
\right)
\left(
\widetilde{W}
\odot
(P-\widetilde{W})^{\top}
\right)
\odot
(P-\widetilde{W})^{\top}
\odot
(P-\widetilde{W})
\right]_{jk}
\\
&=
\frac{1}{2}
L^{\top}
\left[
\left(
\widetilde{W}^{\top}
\odot
(P-\widetilde{W})
\right)
\left(
\widetilde{W}
\odot
(P-\widetilde{W})^{\top}
\right)
\odot
(P-\widetilde{W})^{\top}
\odot
(P-\widetilde{W})
\right]
L .
\end{aligned}
\end{equation}

The counts of the other directed three-node motifs can be obtained analogously by replacing the corresponding edge-present and edge-absent terms according to their motif topologies.

\section{Additional Experiments}

\subsection{Minimum Rank for Applying Motif Constraints}

We further analyze the minimum rank required for representing the motif constraint. 
For each three-node motif, its adjacency pattern can be written as a motif matrix
\begin{equation}
    M = M_I M_J,
\end{equation}
where
\begin{equation}
    M_I \in \mathbb{R}^{n \times r},
    \qquad
    M_J \in \mathbb{R}^{r \times n}.
\end{equation}
Since
\begin{equation}
    \mathrm{rank}(M) \leq r,
\end{equation}
the rank $r$ determines the maximum structural complexity that can be represented by the low-rank motif factorization.

\begin{figure}[htbp]
    \centering
    \includegraphics[width=1\linewidth]{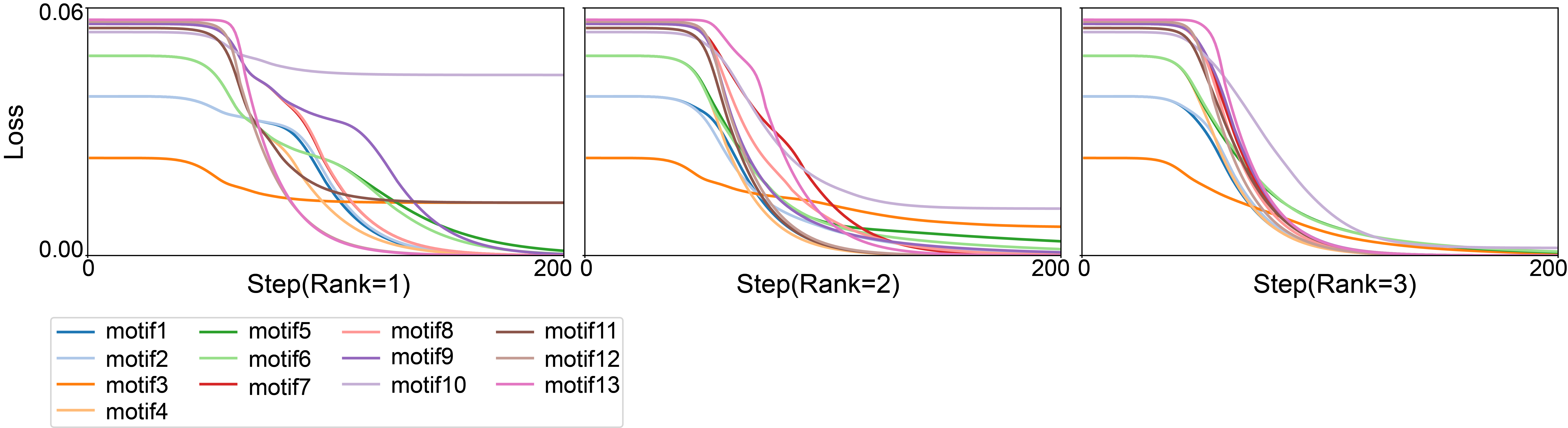}
    \caption{Rank requirement for low-rank motif factorization. 
    Each three-node motif matrix is factorized as $M = M_I M_J$, where $M_I \in \mathbb{R}^{3 \times r}$ and $M_J \in \mathbb{R}^{r \times 3}$. 
    Rank-one factorization is sufficient for most motif constraints, except motif 3 and motif 10. 
    Motif 3 represents a feed-forward directed structure, while motif 10 represents a cyclic directed structure. 
    Increasing the rank to two can effectively support almost all motif constraints considered in this work.}
    \label{fig:min_rank_motif}
\end{figure}

As shown in Figure~\ref{fig:min_rank_motif}, although the motif matrix is defined over three nodes, most motif patterns do not require rank three. 
In practice, rank-one factorization is already sufficient to represent or constrain most motif structures. 
The main exceptions are motif 3 and motif 10. 
Motif 3 corresponds to a feed-forward directed structure, while motif 10 corresponds to a cyclic directed structure. 
These two structures require richer directional interactions than those expressible by a rank-one motif matrix.

When the rank is increased to 2, almost all motif matrices can be effectively represented by the factorization $M = M_I M_J$. 
This observation shows that the motif constraint can be imposed with a very small rank. 
Therefore, Motif-Mamba can incorporate biologically inspired motif priors without relying on a high-rank or dense coupling matrix.

\subsection{Additional Parameter Cost}

Motif-Mamba introduces only a small number of additional parameters compared with the original Mamba architecture. 
The additional parameters mainly come from the motif-constrained low-rank coupling module, including the low-rank motif factors $M_I$ and $M_J$, together with a learnable scalar coefficient $k_{\mathrm{motif}}$. 
Instead of assigning an independent motif coupling matrix to every channel, Motif-Mamba adopts a block-wise sharing strategy.

Specifically, different Mamba blocks use different motif factorization matrices $M_I$ and $M_J$, allowing each block to learn its own motif-guided recurrent structure. 
This design provides sufficient flexibility across network depth, since different layers can capture different levels of temporal and structural interactions. 
However, within the same Mamba block, all channels share the same pair of motif matrices $M_I$ and $M_J$. 
The scalar coefficient $k_{\mathrm{motif}}$ is also assigned at the block level to control the strength of the low-rank motif coupling. 
Therefore, the motif prior is applied consistently across channels in the same block, rather than introducing a separate set of motif parameters for each channel.

For each Mamba block, the additional number of parameters is
\begin{equation}
    2 d_{\mathrm{state}} r + 1,
\end{equation}
where $2 d_{\mathrm{state}} r$ corresponds to the two low-rank motif factors $M_I \in \mathbb{R}^{d_{\mathrm{state}} \times r}$ and $M_J \in \mathbb{R}^{r \times d_{\mathrm{state}}}$, and the additional $1$ corresponds to the learnable scalar coefficient $k_{\mathrm{motif}}$. 
Thus, for a model with $n_{\mathrm{layer}}$ blocks, the total number of additional parameters is
\begin{equation}
    n_{\mathrm{layer}} \left( 2 d_{\mathrm{state}} r + 1 \right).
\end{equation}

This block-wise parameterization greatly reduces the additional parameter cost. 
Because $M_I$ and $M_J$ are shared among channels within each block, the number of newly introduced parameters mainly depends on the number of blocks and the selected low-rank dimension, rather than scaling independently with every channel. 
As a result, Motif-Mamba can introduce biologically inspired motif constraints into the recurrent dynamics while preserving nearly the same model scale as the original Mamba.

An additional advantage of this design is its extensibility. 
Since different blocks maintain independent motif factorization matrices, different motif priors can be assigned to different blocks in future extensions. 
For example, lower blocks may be constrained by simple local interaction motifs, while higher blocks may incorporate more complex recurrent or cyclic motifs. 
This provides a flexible way to study hierarchical motif organization across depth, enabling Motif-Mamba to serve as a general framework for injecting layer-specific biological structural priors into sequence models.

As shown in Table~\ref{tab:additional_param_cost}, the parameter sizes of Motif-Mamba remain very close to those of the corresponding Mamba baselines across different model scales. 
This comparison indicates that the performance improvement of Motif-Mamba is not mainly caused by increasing the number of parameters. 
Instead, the gain comes from introducing motif-guided structural priors through a lightweight block-wise low-rank coupling design.

\begin{table}[t]
\centering
\small
\caption{Additional parameter cost of Motif-Mamba. The added parameters come from the low-rank motif factors $M_I$ and $M_J$, together with a learnable scalar coefficient $k_{\mathrm{motif}}$. Different blocks use different motif factorization matrices, while channels within the same block share the same factors.}
\label{tab:additional_param_cost}
\begin{tabular}{lrrrrrr}
\toprule
\textbf{Model} & 
\textbf{Original Params} & 
$\bm{n_{\mathrm{layer}}}$ &
$\bm{d_{\mathrm{state}}}$ &
\textbf{Rank $r$} & 
\textbf{Added Params} & 
\textbf{Ratio} \\
\midrule
Motif-Mamba-130M & 129,135,360   & 24 & 16 & 2 & 1,560 & 0.001208\% \\
Motif-Mamba-370M & 371,516,416   & 48 & 16 & 2 & 3,120 & 0.000840\% \\
Motif-Mamba-790M & 793,204,224   & 48 & 16 & 2 & 3,120 & 0.000393\% \\
Motif-Mamba-1.4B & 1,372,178,432 & 48 & 16 & 2 & 3,120 & 0.000227\% \\
\bottomrule
\end{tabular}
\end{table}

\subsection{Train Tokens}

We further analyze the training dynamics of Motif-Mamba with respect to the number of training tokens. 
The continued training data are randomly sampled from the Pile dataset, which is the same data source used for training the original Mamba models. 
This setting keeps the data distribution consistent with the Mamba backbone and avoids introducing additional gains from a different training corpus.

\begin{figure}[t]
    \centering
    \includegraphics[width=0.5\linewidth]{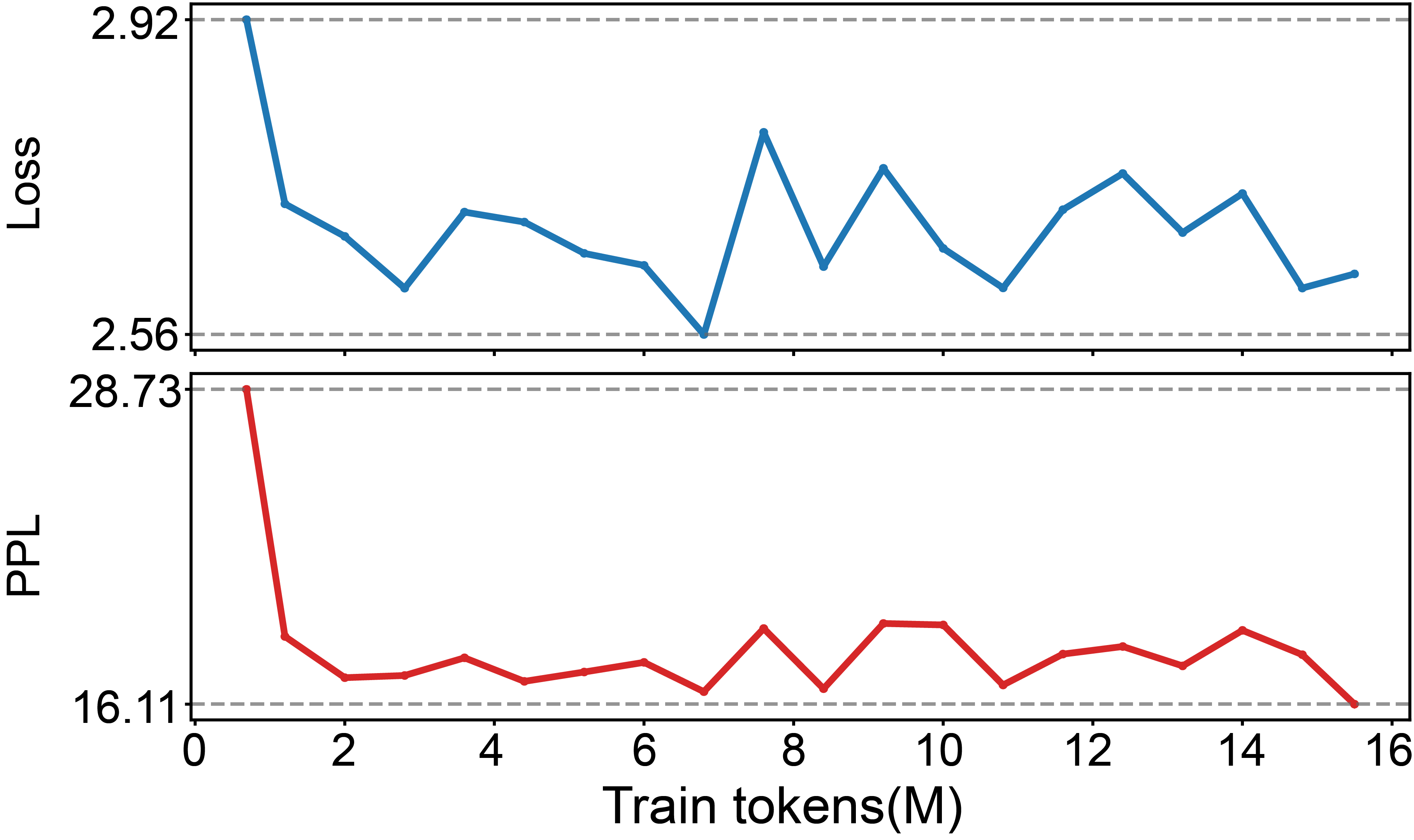}
    \caption{
    Training dynamics of Motif-Mamba with respect to the number of training tokens.
    The continued training data are randomly sampled from the Pile dataset, following the same data source used for the original Mamba backbone.
    The loss and perplexity decrease quickly in the early stage and become relatively stable after about 2M tokens.
    Continued training of the original Mamba backbone under the same token budget does not lead to noticeable improvement, indicating that the performance gain mainly comes from the motif-constrained low-rank recurrent pathway.
    Based on this trend, we train Motif-Mamba with 10M tokens in the main experiments.
    }
    \label{fig:train_tokens}
\end{figure}

Figure~\ref{fig:train_tokens} shows the loss and perplexity curves during training with up to 16M tokens. 
Both loss and PPL decrease rapidly at the early stage of training and become largely stable after approximately 2M tokens. 
Although small fluctuations remain in the later training stage, the overall performance does not show a clear additional improvement with more tokens.

To further exclude the possibility that the observed improvement is mainly caused by continued training, we also continue training the original Mamba backbone under the same token budget and training setting. 
The continued training of Mamba does not bring noticeable performance improvement, suggesting that the gains of Motif-Mamba mainly come from the motif-constrained low-rank recurrent pathway rather than from additional training tokens alone.

Based on this observation, we use 10M training tokens for the main experiments. 
This setting provides a stable training regime while avoiding unnecessary computational cost from longer training.

\section{Experimental Details}
\label{app:experimental_details}

\subsection{Model Variants}

We evaluate several variants of Mamba and Motif-Mamba. Standard Mamba is used as the main baseline. Motif-Mamba extends Mamba by adding a low-rank recurrent coupling term into the state update. To analyze the role of motif constraints, we further evaluate Motifblank-Mamba, which keeps the same low-rank pathway but removes the motif constraint, as well as Motif2-Mamba and Motif12-Mamba, which apply different motif-based structural constraints.

Unless otherwise specified, Motif-Mamba refers to the Motif2-constrained variant in the main experiments. All Motif-Mamba variants are built on the same Mamba backbone as their corresponding baselines, and the remaining architecture is kept unchanged except for the inserted low-rank matrices $M_I$ and $M_J$.

\subsection{Training Setup for Language Evaluation and Motif Ablation}

The training setup in this subsection is used for the language model evaluation and motif ablation experiments. For these experiments, Motif-Mamba is initialized from pretrained Mamba checkpoints and then continuously trained after inserting the low-rank recurrent coupling matrices $M_I$ and $M_J$. This setting allows us to evaluate whether the proposed motif-constrained dynamics can improve the original Mamba backbone without training the whole model from scratch.

All models are trained on NVIDIA A800 GPUs. The context length is fixed to 2048 tokens. We use 10M additional training tokens for the main language evaluation and motif ablation experiments to balance training stability and computational cost.

Table~\ref{tab:training_hyperparams} summarizes the main hyperparameters used for continued training at different model scales. Motif-Mamba models are initialized from pretrained Mamba checkpoints and continuously trained with the inserted low-rank recurrent coupling.

\begin{table}[h!]
\centering
\small
\caption{
Training hyperparameters for language model evaluation and motif ablation.
}
\label{tab:training_hyperparams}
\begin{tabular}{lcccc}
\toprule
\textbf{Model} 
& \textbf{GPUs} 
& \textbf{Train Tokens} 
& \textbf{Batch Size} 
& \textbf{Learning Rate} \\
\midrule
Motif-Mamba-130M & 1$\times$A800 & 10M & 2 & 1e-3 \\
Motif-Mamba-370M & 1$\times$A800 & 10M & 1 & 1e-3 \\
Motif-Mamba-790M & 2$\times$A800 & 10M & 1 & 1e-3 \\
Motif-Mamba-1.4B & 4$\times$A800 & 10M & 1 & 1e-3 \\
\bottomrule
\end{tabular}
\end{table}

\subsection{Long-Sequence Extrapolation}

The long-sequence extrapolation experiment uses a separate training setup from the language model evaluation experiments. In this experiment, models are trained on the induction heads task with a sequence length of 256 and a vocabulary size of 16. The goal is to evaluate whether each model can generalize to contexts longer than those observed during training.

At test time, models are evaluated on sequence lengths ranging from 64 to 16384 tokens. We report next-token prediction accuracy on the induction target positions and visualize performance as the test sequence length increases. All models in this experiment use the same data generation process, optimizer, batch size, learning rate schedule, and number of training steps.

Table~\ref{tab:induction_model_config} summarizes the model configurations used in the long-sequence extrapolation experiment. All models are kept at a comparable scale to ensure a fair comparison across different sequence modeling architectures.

\begin{table}[h!]
\centering
\small
\caption{Model configurations used in the long-sequence extrapolation experiment.}
\label{tab:induction_model_config}
\begin{tabular}{lccc}
\toprule
\textbf{Model} & \textbf{\#Layers} & \textbf{Hidden Dim.} & \textbf{Additional Settings} \\
\midrule
Transformer & 2 & 64 & n\_heads = 4, ffn\_mult = 2.0 \\
LSTM & 2 & 64 & batch\_first = True, no heads \\
RWKV & 2 & 72 & batch\_first = True, no heads \\
Mamba & 2 & 64 & $d_{\mathrm{state}} = 16$, $d_{\mathrm{conv}} = 4$, expand = 2 \\
Motif-Mamba & 2 & 64 & $d_{\mathrm{state}} = 16$, $d_{\mathrm{conv}} = 4$, expand = 2, rank = 2 \\
\bottomrule
\end{tabular}
\end{table}

As a complementary quantitative summary, Table~\ref{tab:induction_heads_extrapolation} reports the accuracy values across different evaluation lengths. Motif-Mamba maintains higher accuracy than the compared baselines as the test sequence length increases, indicating stronger extrapolation ability beyond the training context.

\begin{table}[h!]
\centering
\small
\caption{Accuracy on the induction heads extrapolation task (\%). Models are trained with sequence length 256 and evaluated on different test sequence lengths.}
\label{tab:induction_heads_extrapolation}
\begin{tabular}{lccccccccc}
\toprule
\textbf{Model} & \textbf{64} & \textbf{128} & \textbf{256} & \textbf{512} & \textbf{1024} & \textbf{2048} & \textbf{4096} & \textbf{8192} & \textbf{16384} \\
\midrule
LSTM        & 99.12 & 98.74 & 97.86 & 28.63 & 8.27  & 3.14  & 1.78  & 1.83  & 1.76  \\
Transformer & 99.43 & 99.21 & 98.67 & 34.58 & 10.36 & 4.82  & 2.87  & 2.91  & 2.76  \\
RWKV        & 99.76 & 99.48 & 98.91 & 67.42 & 41.73 & 23.68 & 15.42 & 11.36 & 7.84  \\
Mamba       & 99.82 & 99.61 & 99.08 & 75.64 & 48.72 & 30.86 & 20.74 & 15.28 & 10.93 \\
Motif-Mamba & 99.96 & 99.84 & 99.37 & 81.72 & 55.63 & 35.81 & 23.76 & 17.42 & 11.87 \\
\bottomrule
\end{tabular}
\end{table}

\subsection{BCI Decoding}

For the BCI decoding experiment, we use a separate set of model configurations from the language modeling experiments. The goal is to compare different sequence models under a controlled neural decoding setting. All models are trained and evaluated using the same data split, preprocessing pipeline, and evaluation protocol.

The model input is a multi-channel neural activity sequence, and the target is the corresponding two-dimensional wrist force/kinematic trajectory. We use mean squared error and coefficient of determination as decoding metrics. For single-day evaluation, each recording day is split into training, validation, and test segments. For cross-day generalization, models are trained on sessions before the dashed line and evaluated on held-out future sessions. Error bars are computed across repeated runs with different random seeds.

Table~\ref{tab:bci_model_config} summarizes the model configurations used in the BCI decoding experiment. LSTM and Transformer are included as common sequence modeling baselines. Mamba and Motif-Mamba share the same backbone configuration, while Motif-Mamba additionally introduces motif-constrained low-rank recurrent coupling with rank $r=2$.

\begin{table}[h!]
\centering
\small
\caption{Model configurations used in the BCI decoding experiments. Motif-Mamba and Mamba use the same backbone configuration.}
\label{tab:bci_model_config}
\begin{tabular}{lccc}
\toprule
\textbf{Model} & \textbf{\#Layers} & \textbf{Hidden Dim.} & \textbf{Additional Settings} \\
\midrule
LSTM & 2 & 64 & dropout = 0.2 \\
Transformer & 2 & 64 & n\_heads = 4, dim\_ff = 128, dropout = 0.2 \\
Mamba & 2 & 56 & $d_{\mathrm{state}} = 8$, $d_{\mathrm{conv}} = 4$, expand = 1 \\
Motif-Mamba & 2 & 56 & $d_{\mathrm{state}} = 8$, $d_{\mathrm{conv}} = 4$, expand = 1, rank = 2 \\
\bottomrule
\end{tabular}
\end{table}

\section{Related Work}

\subsection{State Space Models for Sequence Modeling}

State space models (SSMs) have become an important family of efficient sequence models. Unlike attention-based models that explicitly compute pairwise token interactions, SSMs model dependencies through recurrent state evolution and can be implemented efficiently in recurrent or convolutional forms. Representative models such as S4, S5, and H3 have shown strong performance in long-range sequence modeling and language modeling \citep{gu2022s4,smith2023s5,fu2023h3}.

Mamba further advances this direction by introducing selective state space models with input-dependent parameters and a hardware-aware parallel scan algorithm \citep{gu2023mamba}. This design enables linear scaling with sequence length and makes Mamba suitable for efficient long-context modeling. However, Mamba still relies heavily on diagonal or element-wise state transitions, which limits explicit interaction among hidden state dimensions. This motivates our use of structured low-rank dynamics to enhance cross-dimensional interactions while preserving efficiency.

\subsection{Low-Rank Modeling in Deep Learning}

Low-rank decomposition is widely used for model compression, acceleration, and efficient adaptation. By factorizing high-dimensional matrices into products of lower-rank matrices, low-rank methods can reduce parameter and computational costs while maintaining expressive capacity \citep{sainath2013lowrank}. In Transformer-based models, Linformer reduces the complexity of self-attention through low-rank approximation, while LoRA injects trainable low-rank matrices for parameter-efficient adaptation \citep{wang2020linformer,hu2022lora}.

Low-rank structures are also closely related to compact dynamical representations. In recurrent neural networks, low-rank connectivity has been shown to induce low-dimensional dynamics that link network structure, state evolution, and computation \citep{mastrogiuseppe2018linking}. However, most existing low-rank methods are generic and do not explicitly impose structural constraints on the learned interaction. Motif-Mamba addresses this limitation by introducing motif-constrained low-rank coupling into the Mamba state update.

\subsection{Network Motifs and Structured Priors}

Network motifs are recurrent and statistically over-represented connectivity patterns among small groups of nodes in complex networks \citep{milo2002network}. Such local structures have been studied in biological, ecological, and engineered systems. In neuroscience, connectome studies have revealed structured connectivity patterns in the mouse brain, suggesting that local circuit organization may support efficient information processing \citep{oh2014mesoscale,sadovsky2014mouse}.

Motif information has also been used in graph learning to capture higher-order structural patterns beyond pairwise edges. Motif-based graph neural networks have shown that incorporating motif-level information can improve representation learning on graph-structured data \citep{monti2018motifnet,chen2023motif}. However, motif-based structural priors remain underexplored in efficient sequence models such as Mamba. Motif-Mamba integrates motif-constrained low-rank dynamics into selective state space models to enhance structured cross-dimensional interactions while preserving linear-time sequence modeling.
\section{Code Availability}

The source code is available at:\url{https://anonymous.4open.science/r/Motif-Mamba-1227/}

\end{document}